\documentclass[journal]{journal}
\ifCLASSINFOpdf
\else
\fi
\usepackage{graphicx}
\usepackage{caption}
\usepackage{amsmath}
\usepackage{amssymb}
\usepackage{booktabs}
\usepackage{makecell}
\usepackage[pagebackref,breaklinks,colorlinks]{hyperref}
\usepackage{cite}

\usepackage{float}
\usepackage{etoolbox}
\usepackage{dirtytalk}
\usepackage{multirow}
\AtEndEnvironment{thebibliography}
{

\bibitem{eighteen} 
\url{https://github.com/llvll/imgcluster} [Online]. Accessed: Aug. 10, 2021.

}

\begin{document}
%
\title{Automatic Detection and Classification of Symbols in Engineering Drawings}
%
%
%

\author{Sourish~Sarkar,
        Pranav Pandey
        and~Sibsambhu~Kar
\thanks{The authors are with Wipro Limited, corresponding author: Sourish Sarkar, e-mail: sourish.sarkar@wipro.com.}}


\markboth{Journal of \LaTeX\ Class Files,~Vol.~6, No.~1, January~2007}%
{Shell \MakeLowercase{\textit{et al.}}: Bare Demo of IEEEtran.cls for Journals}

\maketitle
\thispagestyle{empty}

\begin{abstract}
 A method of finding and classifying various components and objects in a design diagram, drawing, or planning layout is proposed. The method automatically finds the objects present in a legend table and finds their position, count and related information with the help of multiple deep neural networks. The method is pre-trained on several drawings or design templates to learn the feature set that may help in representing the new templates. For a template not seen before, it does not require any training with template dataset. The proposed method may be useful in multiple industry applications such as design validation, object count, connectivity of components, etc. The method is generic and domain independent. 
\end{abstract}

\begin{IEEEkeywords}
Engineering drawing, deep learning, object detection, object classification.
\end{IEEEkeywords}

%
\IEEEpeerreviewmaketitle

\section{Introduction}
\label{one}
\IEEEPARstart{I}{dentifying} various components in designs and drawings in an automated fashion may reduce significant human effort and time. It may help in automated design verification by finding the number, location, connectivity, correctness of components, etc. 
\par The goal of this work is to design a system, which takes a target image – an engineering drawing as input and detects various objects present in the image. Next, we wish to count and classify the detected symbols of interest by inferring from the legend section of the image. Note that the type and number of symbols that we wish to detect are not exhaustive and can admit a wide range of variation in their appearance. For instance, a fluorescent lamp in an electrical design diagram can be represented using several different symbols in distinct drawing images. Hence, it is not possible to solve the problem by training an object detection model in a conventional manner which assumes the test examples to be from the known classes.
\par The major components of the proposed work are:
\begin{itemize}
\setlength\itemsep{-.3em}
\item Extraction of the table of legends from the design or drawing.
\item Obtaining the template symbols and their corresponding names/class information.
\item Localizing various symbols appearing throughout the drawing.
\item Counting and classifying each of the detected symbols to one of the classes from the table of legends.
\end{itemize}
\begin{figure}[t]
    \centering
    \includegraphics[width=8cm,height=4cm]{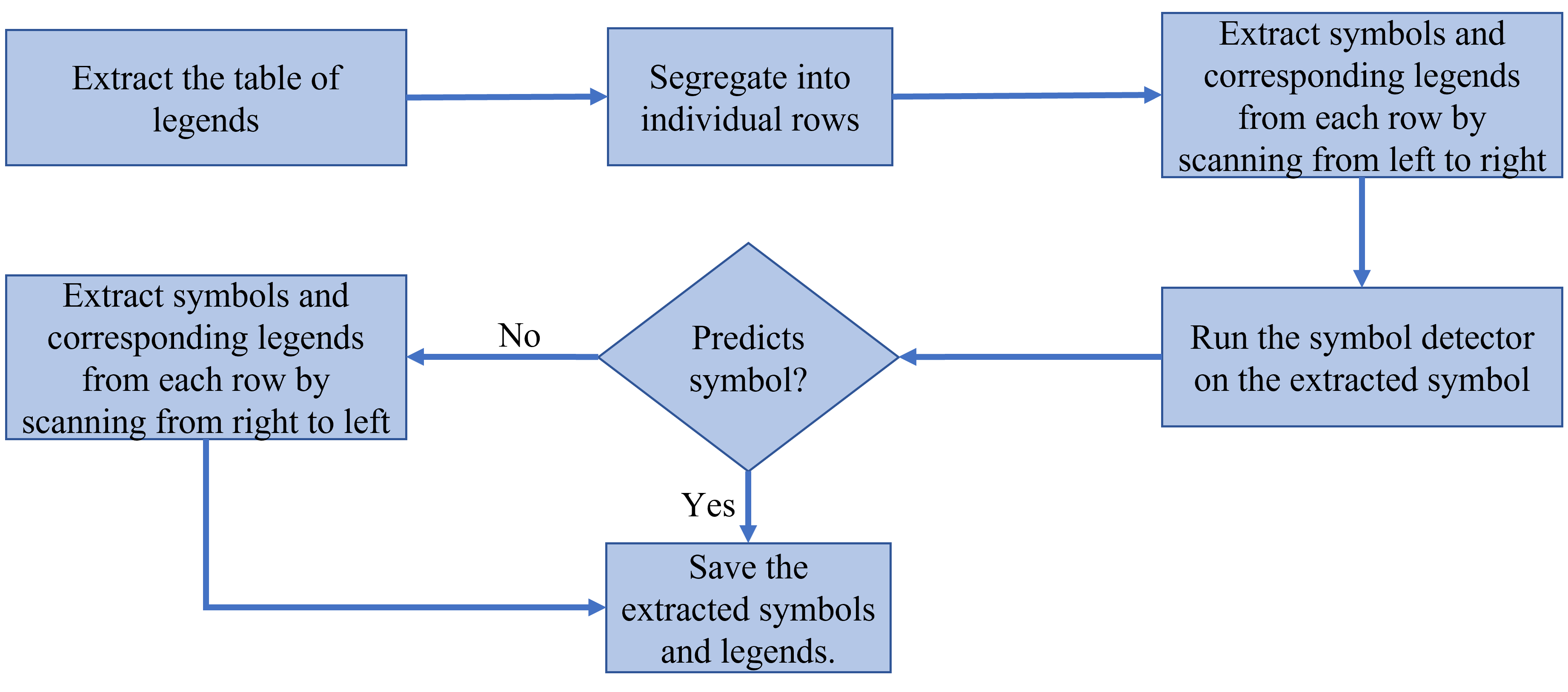}
    \caption{\centering Flowchart of the proposed method for automatic extraction of symbols.}
    \label{f1}
\end{figure}
 A flowchart of the proposed method to extract symbols automatically is provided in Fig. \ref{f1}. The organization of the paper is as follows. A brief review of the literature is provided in Section \ref{one_one}. The various components of the proposed methodology are explained in detail in Section \ref{two}. Section \ref{three} provides a description of the datasets used in this work. The results and discussions are presented in Section \ref{four}. Section \ref{five} concludes the paper.
 
 \begin{figure*}[t]
    \centering
    \includegraphics[width=16cm]{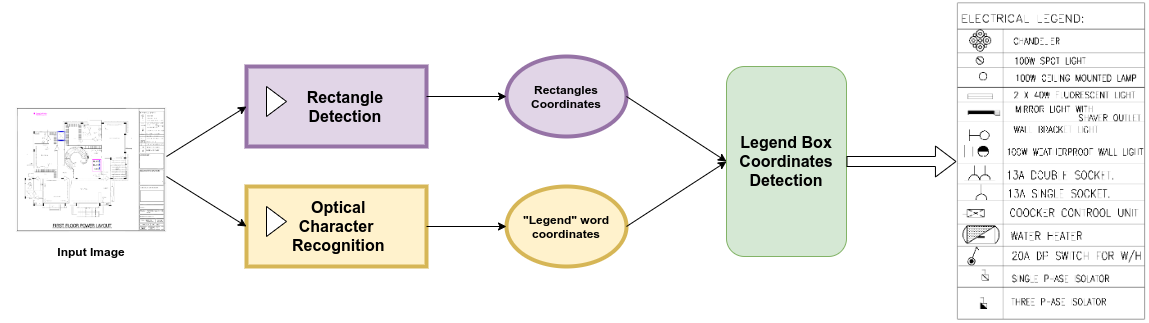}
    \caption{Flowchart of the proposed approach to extract the table of legends.}
    \label{f2}
\end{figure*}
\subsection{Literature survey}
\label{one_one}
The problem of detecting and classifying symbols from engineering drawings (ED) have been addressed in several works, such as \cite{one,two,three,four}. Many of these works were motivated by the problem of digitization of ED wherein the goal is to summarize the relationship between the various symbols. A review of various approaches developed for the purpose of digitizing ED can be found in the work in \cite{one}. Earlier works, such as \cite{five,six} relied on traditional machine learning-based classifiers wherein hand-crafted features were fed to a classifier to recognize symbols.  To circumvent the lack of training data available to train neural networks, the authors in \cite{three} proposed to use a generative adversarial network (GAN) to synthetically generate data for training. Several works in the past decade have addressed the problem of spotting symbols in architectural floor plans. A comprehensive review can be found in \cite{seven}. Recent works have started leveraging deep neural networks for the task of symbol spotting. In \cite{eight}, the authors employed the popular YOLO \cite{nine} model to detect symbols in floor plan diagrams. The task of symbol detection was cast as a semantic segmentation problem in \cite{ten}, thereby attempting to formulate a pixel level approach to detect symbols. The emerging success and popularity of graph neural networks (GNNs) \cite{eleven} have inspired researchers to employ GNNs for the purpose of symbol detection and classification. To this end, the authors in \cite{twelve} proposed to convert a given floorplan image into a region adjacency graph in which every node represented a connected component in the image. A GNN was then used to classify these nodes. The message passing neural network (MPNN) proposed in \cite{thirteen} was employed in \cite{fourteen} to classify symbols. Despite the admittedly rich literature existing in this domain, to the best of our knowledge, the work being reported here is the first attempt of localizing and matching symbols in a zero-shot fashion. 
\section{Proposed method}
\label{two}
Given an engineering diagram, we first aim to extract the table of legends which enlists the various symbols appearing in the drawing and provides a textual description corresponding to each of them. Next, we seek to identify regions in the drawing which possibly contain symbols of interest. Once these regions of interest are identified, the final step is to identify and count the symbols of interest from these regions. The primary challenge in this task is the wide variety of shapes and structures using which these symbols appear in drawings. Further, as mentioned in Section \ref{one}, we cannot expect identical representations of a particular component across multiple drawings. Moreover, the type and variety of templates or objects that are frequented in such diagrams are huge in number. Hence it is impossible to resort to a completely supervised approach, training exhaustively to identify and classify every single type of object that might be encountered in such images. 
\par In an engineering drawing, the information regarding the symbols appearing in the drawing are available in various forms, such as:
\begin{itemize}
\setlength\itemsep{-.3em}
\item A table of legends containing template symbols and the name of the component they depict.
\item A table containing numbers, signifying the index of a component and the corresponding name of the item it depicts.
\item No tabular information associating object names with their corresponding templates.
\end{itemize}
\par In the current work we address the first type of drawing in which the template image is given along with associated component name. In the following sections, we explain in detail the various components of the proposed framework. 
\begin{figure*}[t]
\centering
\begin{tabular}{ccc}
\includegraphics[width=5cm]{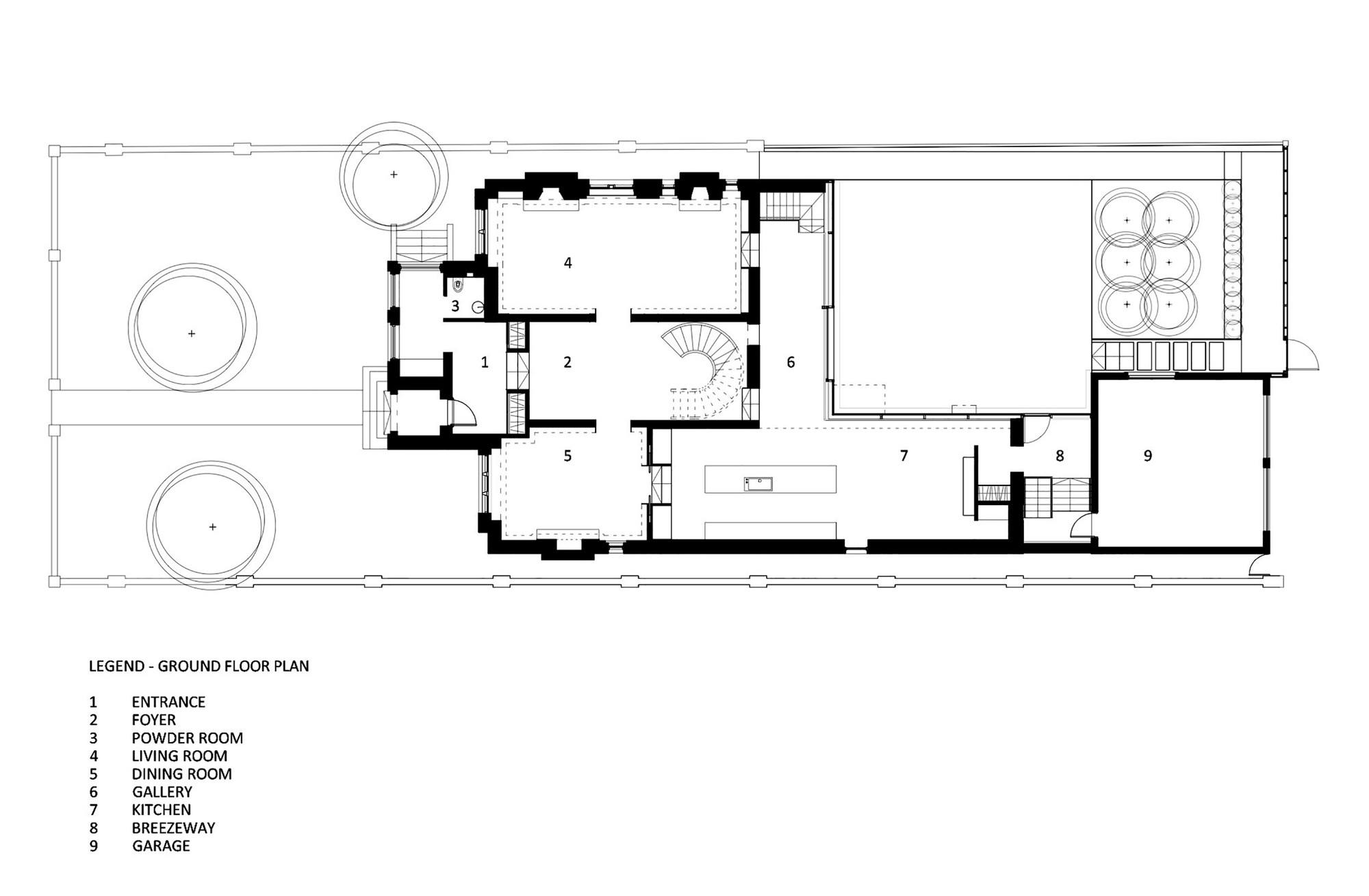}&
\includegraphics[width=5cm]{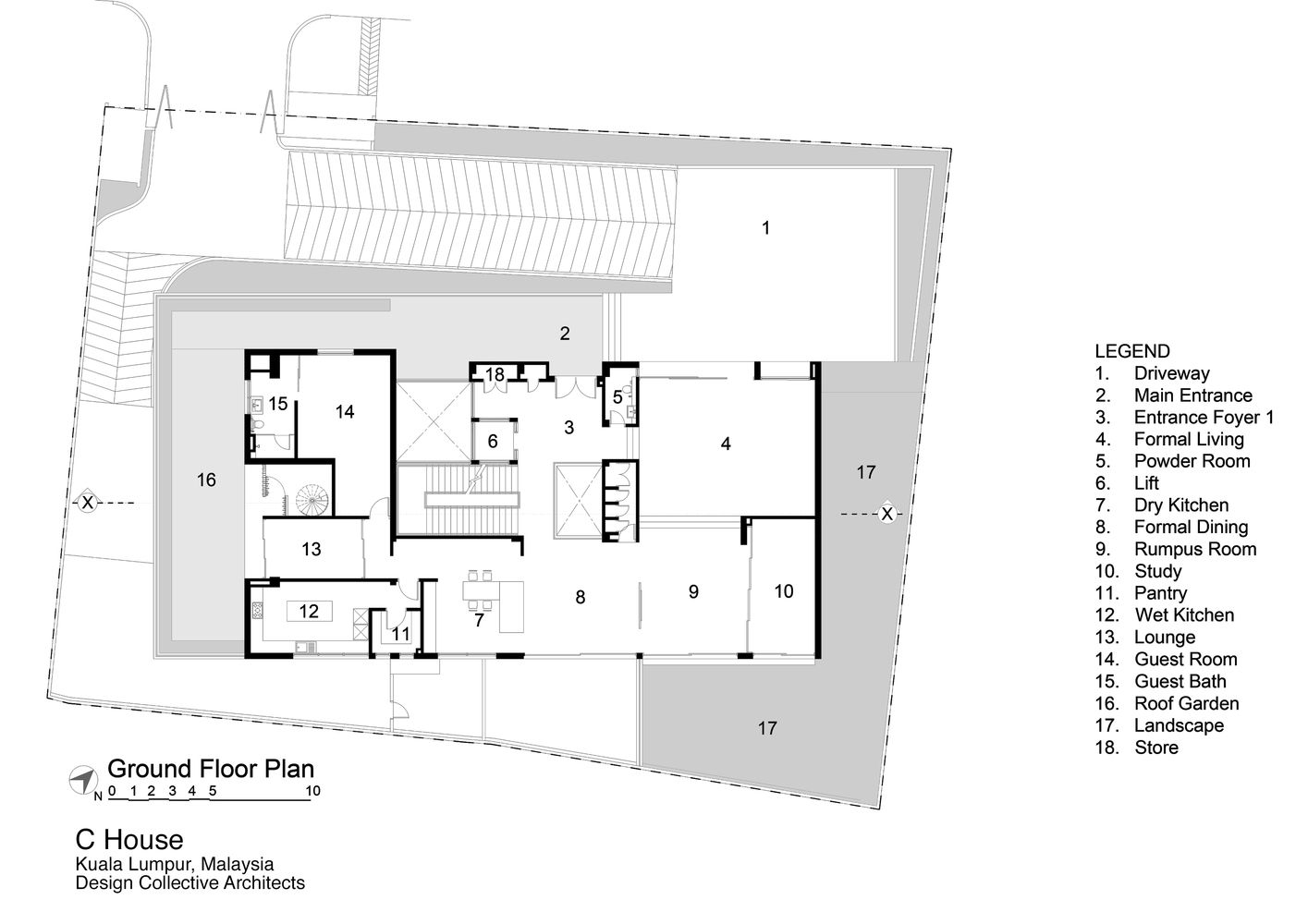}&
\includegraphics[width=5cm]{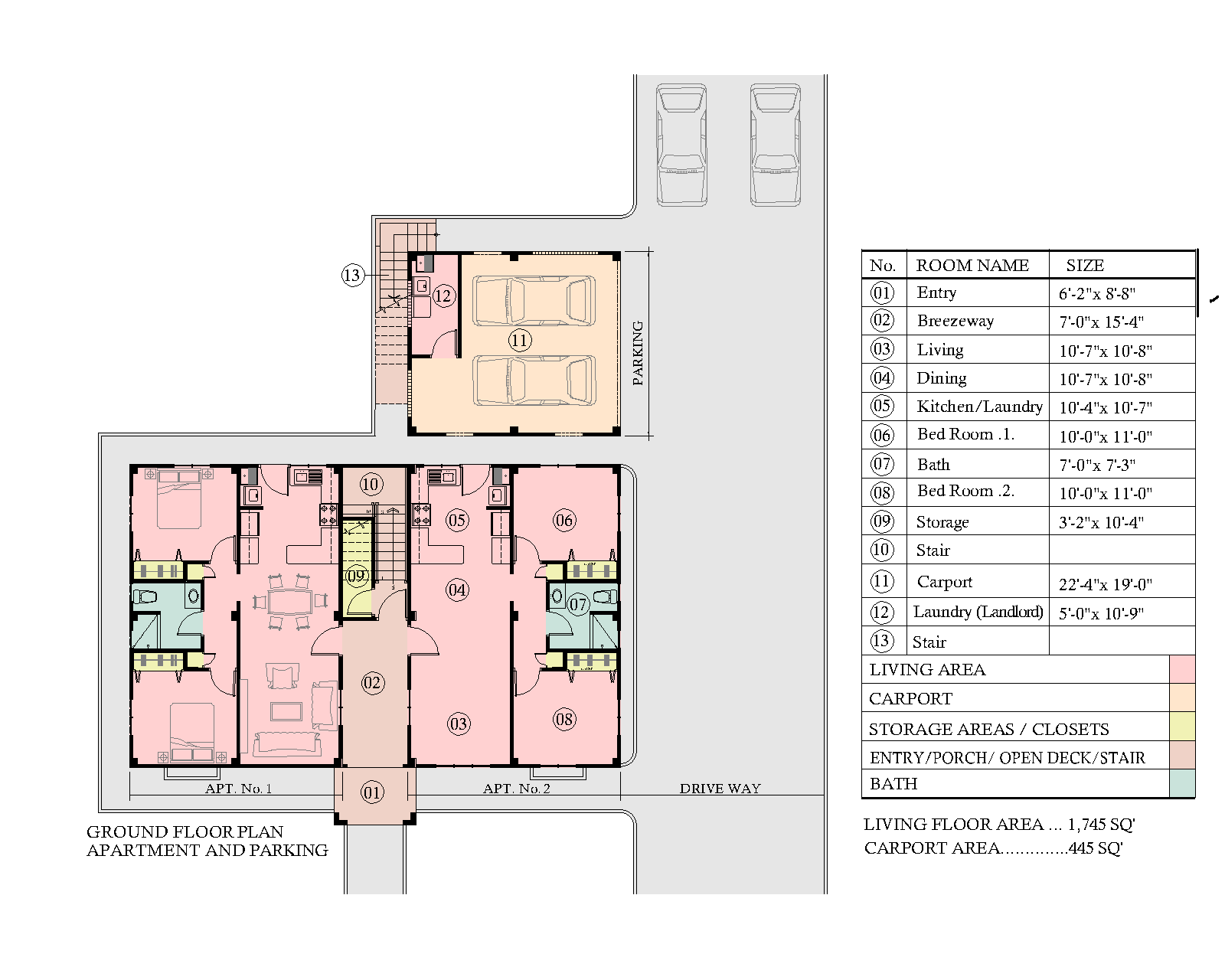}\\
(a) & (b) & (c)\\
\includegraphics[width=5cm]{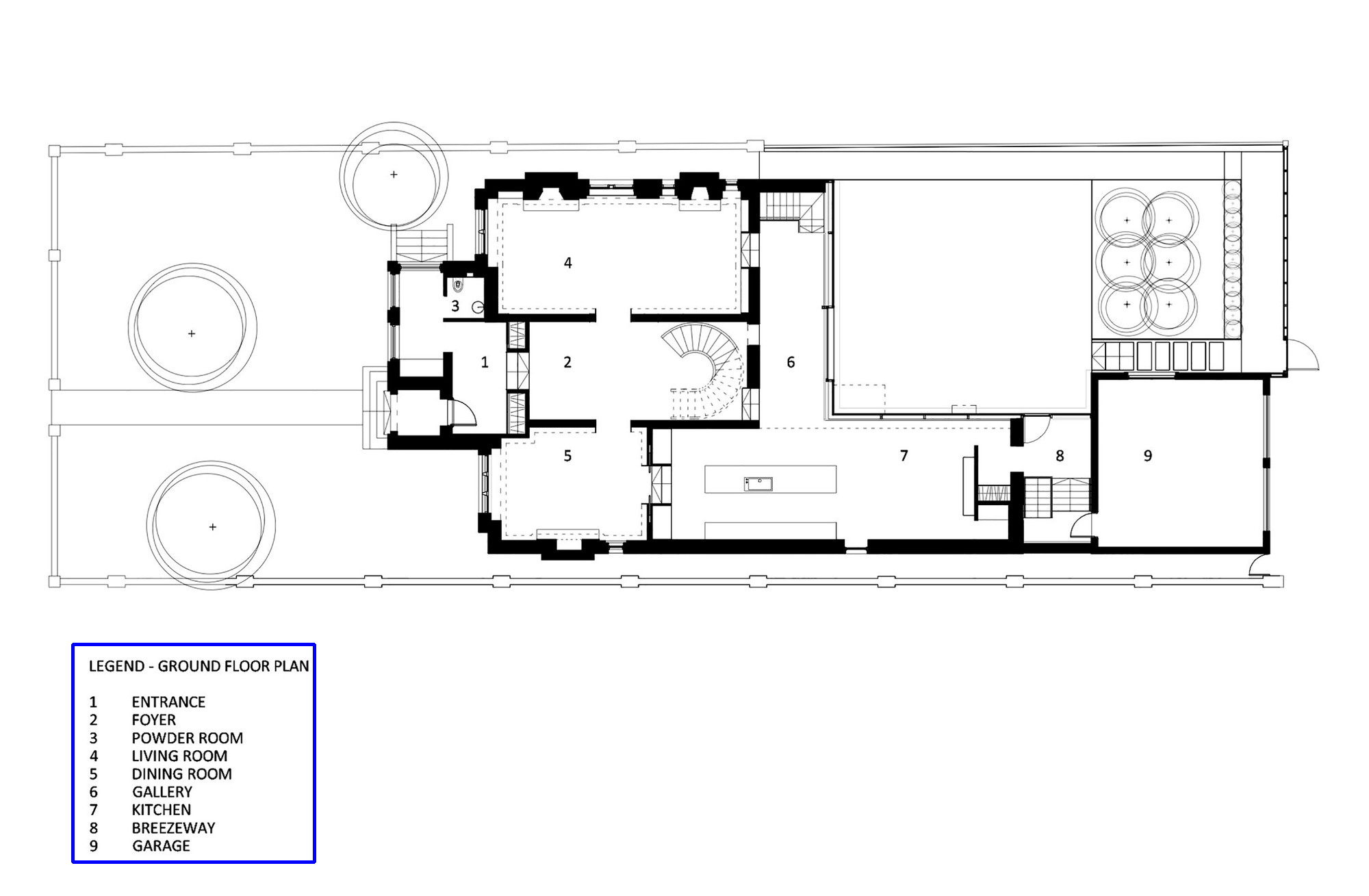}&
\includegraphics[width=5cm]{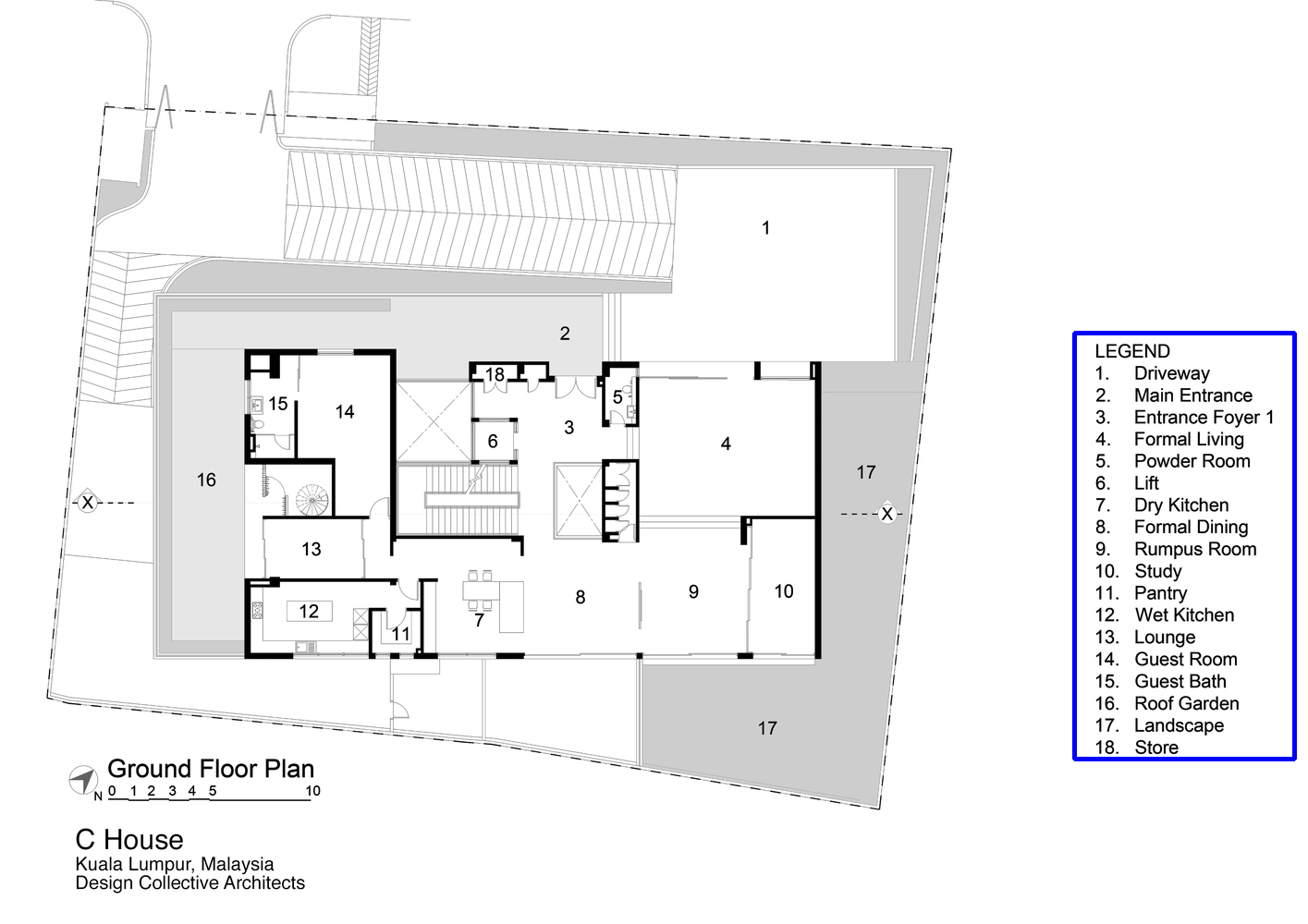}&
\includegraphics[width=5cm]{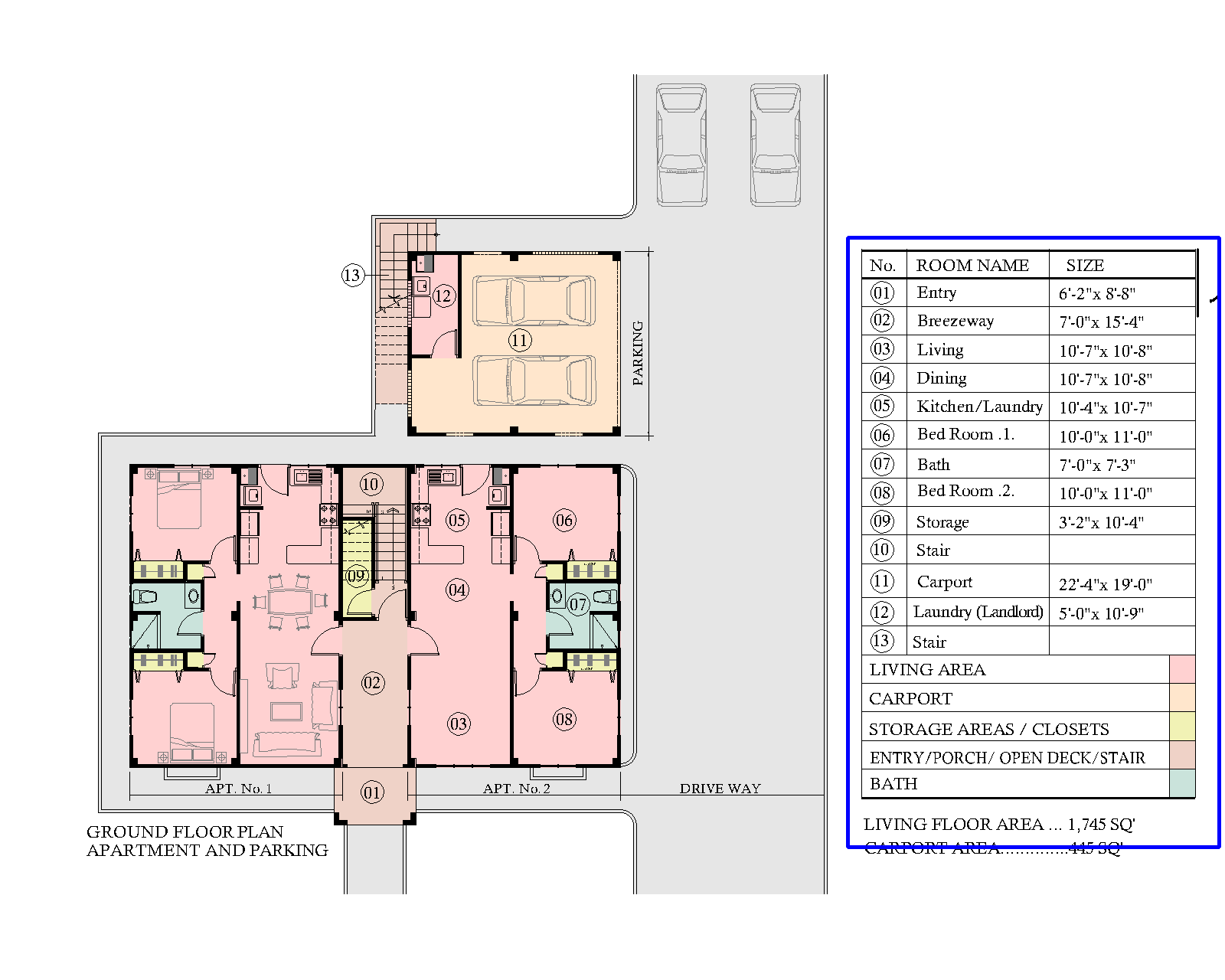}\\
(d) & (e) & (f)\\
\end{tabular}
\caption{\centering Images (a) through (c) show drawings with various forms of tables with legend. Images (d) through (f) show the results of detecting the tables.}
\label{f3}
\end{figure*}
\subsection{Extracting the table of legends}
\label{two_one}
Given an input image, the first task is to identify the information/symbols of interest in the image. To this end, we aim to extract the table of legends in the image which contains information regarding the symbols of interest and their corresponding names/class information. Note that the table may or may not contain lines and borders. Further, the table can be located anywhere in the drawing. This makes the task of detecting the table of legends difficult. To address this task, we adopted two different approaches, which we describe in detail in this section.
\par The first approach involves following multiple preprocessing steps to find the location of the legend box. For an image of width, $w$ and height, $h$, we start by first creating horizontal and vertical kernels of size $1 \times w/70$  and $h/70 \times 1$, respectively, which we use to detect horizontal and vertical lines in the input image. The steps for finding the legend box are as follows.
\begin{itemize}
\setlength\itemsep{-.3em}
    \item We first detect horizontal and vertical lines in the image using the kernels. We apply multiple iterations of morphological erosion and dilation to detect the lines and remove all the unnecessary noise in the image.
    \item We then add both the detected horizontal and vertical lines together to get an image containing all boxes.
    \item Next, we run contour detection on the image and find all the rectangles in the image.
    \item To get the exact location of the legend box, we first run optical character recognition (OCR) on the input image and find the location of the word “LEGEND” in the image.
    \item Once we have the location of the word “LEGEND”, we check which rectangle in the image contains that word inside of it and then crop out the parent rectangle enclosing it.
    \item The extracted parent rectangle is our final output.
\end{itemize}
\par The flowchart in Fig. \ref{f2} shows the complete process for legend box extraction. 
\par Evidently, the above-mentioned approach had several drawbacks. The primary drawback is that the method will not work for tables that do not have borders. Further, the presence of the word ‘LEGEND’ in the table is not guaranteed either. Hence, we next sought to develop a method generic enough to account for the variability in the appearance of these tables.
\par To this end, motivated by the recent success of deep learning-based object detection models, we train a Faster R-CNN \cite{fifteen} model to localize the table of legends present in the drawing. In order to do so, we first collected 51 images of engineering drawings with such tables of legends occurring in various shapes and positions. We then manually annotated these images to create a dataset for training the Faster R-CNN \cite{fifteen} model using transfer learning. We then augmented the data by adding transformations, such as horizontal and vertical flipping, 90 degrees clockwise and counterclockwise rotation and shearing. Fig. \ref{f3} shows some of the various forms of tables that can be encountered in such images and illustrates the performance of the trained object detection model on these images. 
\subsection{Extraction of legends}
\label{two_two}
After extracting the table of legends, the next task is to extract individual legends along with their corresponding names from the table of legends. To do so, we first seek to extract the rows from the table. Each such row contains a symbol and its corresponding name. Once the rows are extracted successfully, the symbol and its corresponding legend are separated. We use simple image processing techniques to accomplish these tasks which we explain in the following paragraphs.
\par To extract the rows, we first binarize the extracted image of the table of legends. The binary image now contains black pixels which correspond to the symbols and legends while the white pixels in the image correspond to the background. We then scan the image from top to bottom and compute the average pixel value of each row. This allows us to detect the rows which contain only white pixels (background) and the rows containing black pixels. This simple approach also equips us with the knowledge about the rows at which a transition occurs from white to black and vice-versa. Hence, we scan the image from top to bottom and extract the rows by cropping sub-regions from one non-white row to the next. Note that such a technique is prone to errors due to stray lines and borders that might be present in the extracted table of legends. To alleviate this problem, as a pre-processing step, we search for rows and columns, more than 90 per cent of pixels of which are black. These rows and columns are then turned white (by assigning a pixel value of 255). 
\begin{figure}[h]
\centering
\begin{tabular}{|c|c|c|}
\hline
Extracted Table & Symbols & Legends\\
\hline
\hspace{0mm} & \hspace{0mm} & \hspace{0mm}\\
\multirow{3}{*}{\includegraphics[scale=0.25]{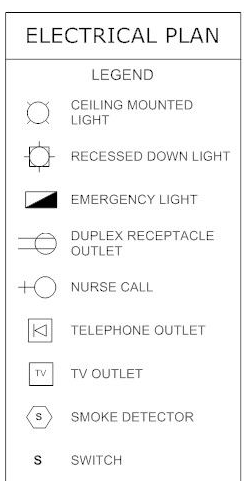}} & \includegraphics[scale=0.5]{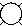} & \includegraphics[scale=0.5]{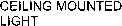}\\
& \includegraphics[scale=0.5]{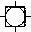} & \includegraphics[scale=0.5]{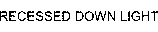}\\
& \includegraphics[scale=0.5]{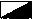} & \includegraphics[scale=0.5]{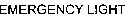}\\
& \includegraphics[scale=0.5]{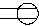} & \includegraphics[scale=0.5]{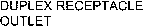}\\
& \includegraphics[scale=0.5]{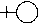} & \includegraphics[scale=0.5]{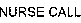}\\
& \includegraphics[scale=0.5]{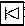} & \includegraphics[scale=0.5]{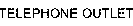}\\
& \includegraphics[scale=0.5]{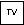} & \includegraphics[scale=0.5]{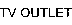}\\
& \includegraphics[scale=0.5]{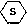} & \includegraphics[scale=0.5]{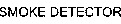}\\
& \includegraphics[scale=0.5]{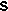} & \includegraphics[scale=0.5]{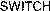}\\
\hspace{0mm} & \hspace{0mm} & \hspace{0mm}\\
\hline
\end{tabular}
    \caption{\centering Detected symbols and legends from the extracted table of legends.}
    \label{f4}
\end{figure}
\par After extracting the rows of the table, the next task is to segregate the symbols and their legends from each of the extracted rows. We use an approach similar to that used for the extraction of rows to perform this task. In this case, however, we scan from left to right instead of scanning from top to bottom and utilize the average pixel value of the columns instead of the rows. Using such an approach, we obtain the symbols and the legends as distinct sub-regions. Note that an issue with the above-mentioned approach is that some rows contain only text (for example the heading of the table). Such rows generally occur towards the top of the table. Eliminating these rows prior to the segregation of the symbols and legends is necessary since these rows would lead to irrelevant and incorrect sub-regions being marked as symbols. We alleviate this problem by using the object detector proposed in Section \ref{two_one} on the extracted rows to search for symbols. We discard the first few rows (if any such rows are present in the table) where the object detector does not predict any symbols.
\par Further, we assume that the symbols appear towards the left of the extracted row while the legends are found towards the right of the symbol. While we did not find many drawings which violate this assumption, this assumption still appears to be restrictive. To circumvent this, we leverage the same object detector proposed in Section \ref{two_one}. We use it to detect objects in the detected sub-region containing the extracted symbol. If no object is detected, we infer that the sub-region must be containing some word from the legend. Hence, the symbol must appear towards the right of the extracted row. We then start scanning from right to left to extract the symbol and the corresponding legend. 
\par Despite being simple, we find this approach capable of extracting the symbols and the legends from tables with varied shape and structure. Fig. \ref{f4} shows extracted symbols and legends using the proposed approach.
\subsection{Localization of symbols}
\label{two_three}
The next step of our framework is to identify regions in the drawing which have a high probability of containing a symbol. Akin to the task of detecting the table of legends, we cast this as an object detection problem wherein we wish to localize the symbols of interest appearing in the image. Note that apart from the background, there is only one class of object to be detected in this problem, namely the symbols of interest. The primary objective here is to train a model which can learn to localize the symbols and ignore the lines, arrows and other stray markings frequented in an engineering drawing. In order to accomplish this task, we resort to the Faster R-CNN \cite{fifteen} model once again. To this end, we first prepare a dataset by cropping regions from several drawings which contain symbols of interest and annotate them. We collected 331 such images and augmented them on the fly while training the model using random horizontal flips. We then trained a Faster R-CNN \cite{fifteen} model pre-trained on the MS-COCO \cite{sixteen} dataset using transfer learning. Fig. \ref{f6} shows the detected symbols using the trained model.
\subsection{Classification of detected symbols}
\label{two_four}
\begin{figure}[t]
    \centering
\begin{tabular}{cc}
\includegraphics[scale = 0.7]{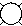}     &
\includegraphics[scale = 0.7]{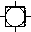}  \\
(a) & (b)\\
\includegraphics[scale = 0.7]{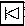}     &
\includegraphics[scale = 0.7]{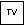}  \\
(c) & (d)
\end{tabular}
    \caption{ Two pairs of symbols extracted from the table of legends that look similar and hence are difficult to distinguish. Note that although it is quite easy for a human to distinguish between (c) and (d), the square boundaries present in both the symbols make it challenging for a classifier to differentiate between them. }
    \label{f5}
\end{figure}

\begin{figure}[h]
    \centering
\begin{tabular}{cc}
\includegraphics[scale = 1]{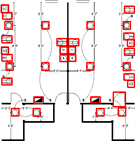}&
\includegraphics[scale = 1]{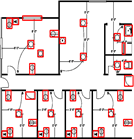}\\
     (a) & (b) 
\end{tabular}
    \caption{\centering Detected symbols in image of drawings using the trained object detection model. }
    \label{f6}
\end{figure}

The final step of the proposed framework is to classify the detected symbols to one of the symbols extracted from the table of legends. We cast this problem as an image similarity measurement problem. That is to say, given a detected symbol, we seek to retrieve the symbol from among the symbols extracted from the table of legends, such that the retrieved symbol is most similar to the query symbol. The number of symbols belonging to a certain class can then be counted as per requirement. However, as mentioned earlier, there could be indefinitely several kinds of symbols appearing in the drawing. Hence it is not feasible to train an image classifier in a conventional supervised fashion. A relatively simple and tempting approach to perform this task would be to employ image similarity metrics such as structural similarity index, peak signal to noise ratio or even the mean squared error between the images. The test symbol could then be classified to the template symbol with which the similarity score is highest (or the mean squared error is the lowest). However, we found that such approaches were not able to distinguish between symbols that look very similar and result in poor classification performance. Fig. \ref{f5} shows an example of such confusing symbols. We therefore seek a method which extracts discriminating features from the query and template images. The best match could then be found using some metric derived from these features that are indicative of the similarity between the images. Further, we wish the extracted features to be robust to rotation, scale, and translation.
\par We use the scale invariant feature transform (SIFT) feature descriptors \cite{seventeen} to extract scale and rotation invariant features from the images. These descriptors are then used to construct a metric using the method used in \cite{eighteen}. This metric indicates the similarity between the two images $I_1$ and $I_2$ containing symbols. Note that, generally, these images have very small dimensions (less than 64 $\times$ 64 pixels). The images $I_1$ and $I_2$ are first resized to 64 $\times$ 64 pixels. Next, the SIFT \cite{seventeen} keypoints and descriptors, ${k_1,d_1}$ and ${k_2,d_2}$, respectively, are extracted from $I_1$ and $I_2$. The extracted keypoints are then matched in a brute-force fashion, comparing every descriptor from ${d_1}$ against every other descriptor from ${d_2}$. Let us denote the total number of matches obtained by $m$. Two keypoints are returned corresponding to each of the $m$ matches. The ratio test as described in \cite{seventeen} is then used to retain only $n$ matches that pass the test. The similarity score $s$ is then calculated as follows:
\begin{equation}
    s = 
    \begin{cases}
    1 & ,n=m\\
    0.1 & ,n=1\\
    1-\dfrac{n}{m} & ,1<n<m\\
    0 & ,\text{otherwise}
    \end{cases}
\end{equation}
We use this similarity score to retrieve the template image which produces the highest score upon comparison with the query image.
\section{Datasets}
\label{three}
Two datasets were created to train the object detection models as mentioned in section \ref{two_one} and section \ref{two_three}. We first collected 51 images to train an object detection model for detecting the tables of legend from engineering drawings. These images contained tables of legends occurring in various shapes and positions. We then manually annotated these images to create a dataset for training the Faster R-CNN \cite{fifteen} model using transfer learning. We then augmented the dataset by adding transformations, such as horizontal and vertical flipping, 90 degrees clockwise and counterclockwise rotation and shearing. Next, we created a dataset for training an object detection model to predict symbols in a given drawing. To this end we cropped regions from several drawings which contained symbols of interest and annotated them. We collected 331 such images and augmented them on the fly while training the model using random horizontal flips. 
\section{Results}
\label{four}
In this section, we assess the performance of the components of the proposed method. Specifically, we aim to assess the performance of the symbol detector and the symbol classifier, explained in sections 2.3 and 2.4, respectively. To do so, we select 11 images of engineering drawing and use the symbol detector to identify regions possibly containing symbols of interest. We then seek to classify these extracted symbols into one of the symbols from the extracted table of legends. Note that several detected symbols do not belong to any of the symbols extracted from the table of legends. This might happen when the detected symbol represents a commonly occurring symbol (such as doors and lamps) whose representation is often omitted from the table of legends. In case of such symbols, we consider the classification to be correct if the classifier fails to associate the symbol to any of the reference symbols and hence labels it as an outlier. We manually check the number of symbols detected and correctly classified and report the results in this section in order to demonstrate the efficacy of the proposed framework. The results are tabulated in Table \ref{t1}. The second column of the table indicates the number of relevant symbols present in the image. The total number of symbols detected by the symbol detector is provided in the third column of the table. Note that this also includes several detected outliers which are not present in the table of legends. The fourth column of the table shows the number of relevant symbols correctly detected while the last column contains the number of symbols classified correctly. From Table \ref{t1}, we find that while 227 relevant symbols are successfully detected out of 301 symbols, the classifier correctly classifies 264 out of the total 354 detected symbols (including outliers).

\begin{table}[h]
    \centering
        \caption{\centering Performance of the symbol detector and the symbol classifier. }
    \begin{tabular}{|c|c|c|c|c|}
    \hline
\makecell{Image\\no.} & \makecell{Number\\ of\\symbols\\ present} & \makecell{Total\\ number\\ of\\ symbols\\ detected} & \makecell{Number\\ of \\symbols\\ detected \\correctly}& \makecell{Number\\ of\\symbols\\ classified\\correctly}\\
\hline
1&33&35&33&31\\
\hline
2&30&27&26&17\\
\hline
3&37&35&29&30\\
\hline
4&24&29&19&26\\
\hline
5&18&20&18&19\\
\hline
6&18&10&9&5\\
\hline
7&32&45&22&30\\
\hline
8&19&34&19&25\\
\hline
9&39&36&19&20\\
\hline
10&13&38&6&31\\
\hline
11&38&45&27&30\\
\hline
Total&301&354&227&264\\
\hline
    \end{tabular}
    \label{t1}
\end{table}
\subsection{Discussions}
\label{four_one}
In this section, we first discuss an alternative approach for the classification of the detected symbols. We then discuss about the drawbacks of the proposed method and also mention our future plans for improving the approach. 
\subsubsection{Siamese network for classification of detected symbols}
\label{four_one_one}
We use a custom designed CNN (Convolutional Neural Network) in a Siamese network \cite{nineteen} setting to train the model to extract discriminating features. We learn image/symbol representations via a supervised metric-based approach with Siamese neural networks \cite{nineteen} wherein we learn the representation by using a Triplet Loss \cite{twenty}, and then reuse that network’s features for recognition without any retraining.
\par A Siamese Network consists of twin networks which accept distinct inputs but are joined by an energy function at the top. This function computes a metric between the highest-level feature representation on each side. The parameters between the twin networks are tied. Weight tying guarantees that two extremely similar images are not mapped by each network to very different locations in feature space because each network computes the same function. The network is symmetric, so that whenever we present two distinct images to the twin networks, the top conjoining layer will compute the same metric as if we were to present the same two images but to the opposite twins. Intuitively, instead of trying to classify inputs, a Siamese network learns to differentiate between inputs, learning their similarity.
\par
\begin{figure}[t]
\centering
\includegraphics[width = 8cm]{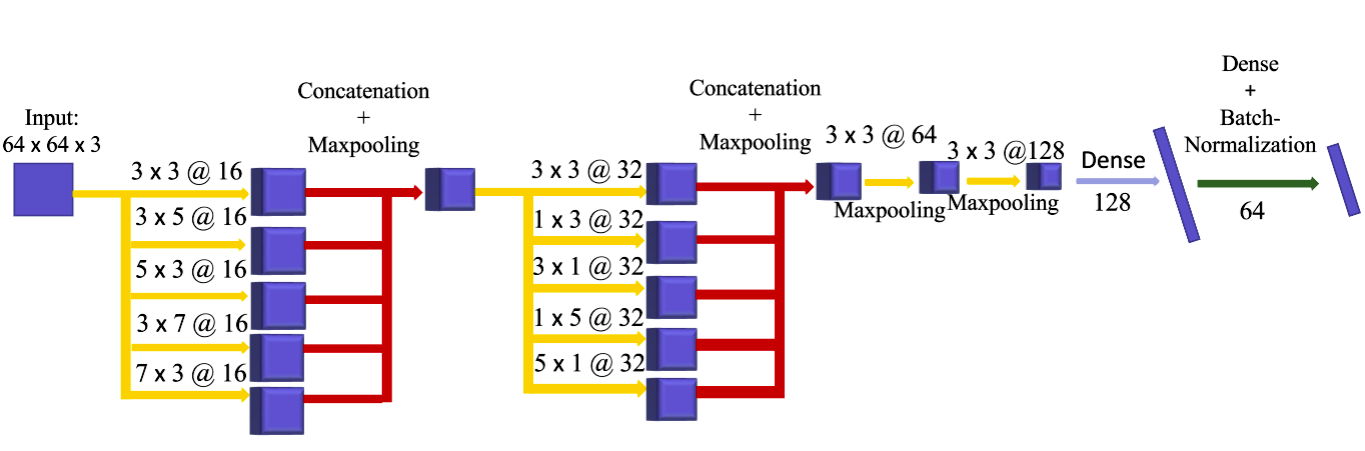}
\caption{Architecture of the proposed network. }
\label{f7}
\end{figure}

 The proposed Siamese network consists of four convolutional layers. In the first two convolutional layers, we adopt a multi-scale approach and carry out convolutions using 16 kernels, each having dimensions of $3 \times 3$, $3 \times 5$, $5 \times 3$, $3 \times 7$, $7 \times 3$ in the first convolutional layer and 32 kernels each of size $3 \times 3$, $1 \times 3$, $3 \times 1$, $3 \times 5$ and $5 \times 3$ in the second layer. We pad the inputs in order to maintain the dimensions of the generated feature maps same as the inputs to the convolutional layers. The feature maps are then concatenated and are followed by a Maxpooling (MP) layer which reduces their spatial dimensions to half as compared to the spatial dimensions of the input tensor. The feature maps obtained from the third and the fourth convolutional layers are also subjected to the MP operation which finally produces a spatial size of $8 \times 8$ after the fourth convolutional layer. The convolutional layers are followed by two fully connected layers. We also apply a batch normalization layer after the second convolutional layer and use ReLU \cite{twentyone} as the activation function in each of the convolutional and the fully connected layers. Fig. \ref{f7} provides a schematic representation of the proposed architecture. 
\par
\begin{figure}[t]
    \centering
    \includegraphics[width=8cm]{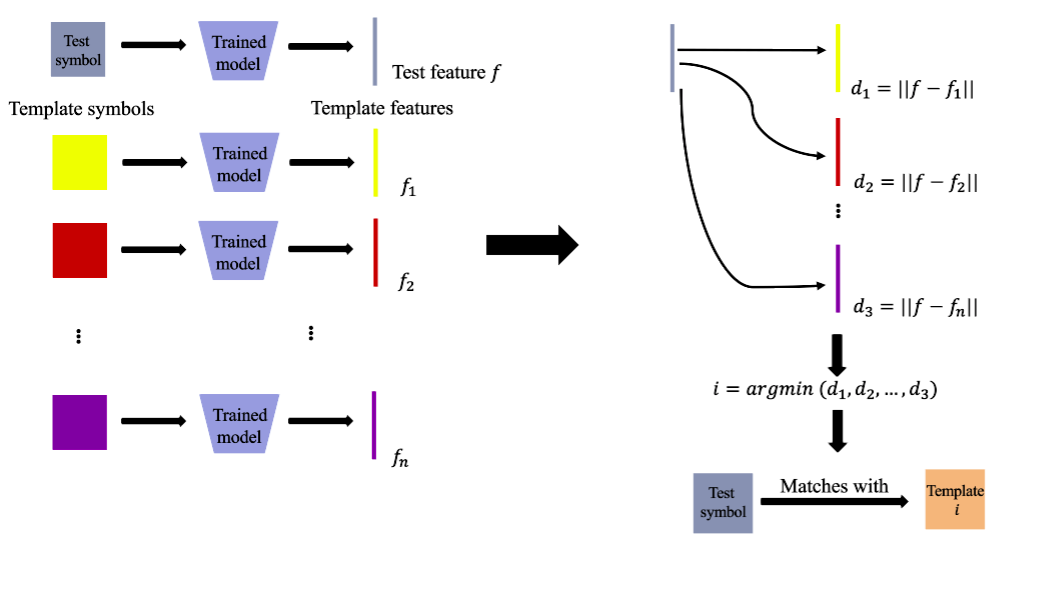}
    \caption{\centering Classifying the detected symbols using the proposed network as a feature extractor.}
    \label{f8}
\end{figure}

 To classify a detected symbol to one of the template symbols, the proposed network is used as a feature extractor. These features are used to assign the test symbols to one of the template symbols in a zero shot fashion. Note that we do not make any assumptions regarding the symbols belonging to one of the classes on which the network was trained. Specifically, by feeding in the test image as input to the proposed network, a 64-dimensional feature vector $\mathbf{f}$ is extracted which serves as the descriptor for the input image. The template images are also presented to the network in order to extract descriptors $\mathbf{f}_1,\mathbf{f}_2,\dots,\mathbf{f}_n$ for each of them. Next, the Euclidean distance between the feature vectors extracted from the test image and the template images is computed as $\mathbf{d}_i=\|\mathbf{f}-\mathbf{f}_i\|, i=1,2,\dots,n$. The test image is then classified to the class of the template image with which the computed distance is minimum. Fig. \ref{f8} provides a graphical description of the proposed framework for inference.
\begin{table}[t]
    \centering
        \caption{\centering Performance of the symbol detector and the symbol classifier using the proposed network as feature extractor.}
    \begin{tabular}{|c|c|c|c|c|}
    \hline
\makecell{Image\\no.} & \makecell{Number\\ of\\symbols\\ present} & \makecell{Total\\ number\\ of\\ symbols\\ detected} & \makecell{Number\\ of \\symbols\\ detected \\correctly}& \makecell{Number\\ of\\symbols\\ classified\\correctly}\\
1&33&35&33&26\\
\hline
2&30&27&26&3\\
\hline
3&37&35&29&24\\
\hline
4&24&29&19&15\\
\hline
5&18&20&18&12\\
\hline
6&18&10&9&2\\
\hline
7&32&45&22&12\\
\hline
8&19&34&19&7\\
\hline
9&39&36&19&19\\
\hline
10&13&38&6&27\\
\hline
11&38&45&27&25\\
\hline
Total&301&354&227&172\\
\hline
    \end{tabular}
    \label{t2}
\end{table}
\par In order to assess the performance of the proposed network, we repeat the experimental procedure described in section \ref{four}. However, this time, we use the proposed network to extract features from the detected symbols. These features are then used to classify the symbol to one of the template symbols as explained in the previous paragraph. Table \ref{t2} shows the performance of the proposed network in classifying the detected symbols from which it can be observed that 172 symbols were classified correctly. 
\par The performance of the proposed network in obtaining discriminating features to classify the detected symbols is inferior as compared to the approach described in section \ref{two_four}. Hence, we chose the former approach for the classification of detected symbols.

\subsubsection{Drawbacks of the proposed approach}
\label{four_one_two}
The primary bottleneck of the proposed approach is the symbol detector mentioned in section \ref{two_three}. This is because the subsequent step, which involves the classification of the detected symbols, depends heavily on the performance of the symbol detector. Although the proposed symbol detector was trained on a variety of symbols frequented in engineering drawings, it was not possible to exhaustively include every kind of symbol in the training set. Fig. 9 shows an example image where several symbols went undetected. We wish to alleviate this drawback in the future work as described in the following sub-section.

\begin{figure}[h]
    \centering
    \includegraphics{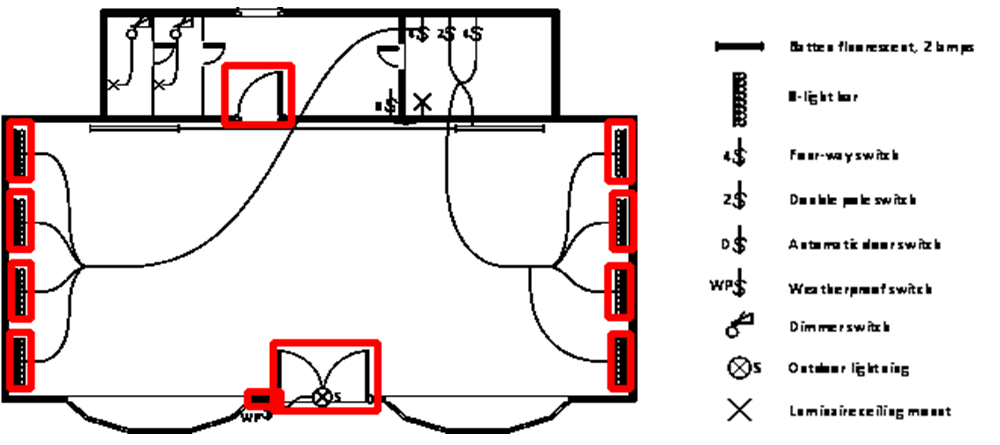}
    \caption{Symbols remaining undetected by the symbol detector: The symbol detector fails to detect the symbols appearing towards the top of the image.}
    \label{f9}
\end{figure}

\subsubsection{Future work}
\label{four_one_three}
As mentioned in section \ref{four_one_one}, the step involving the classification of symbols depends heavily on the performance of the symbol detector. However, in future, we wish to eliminate the symbol detector from the framework. Once the reference symbols are extracted from the table of legends, we aim to search for each of the extracted symbols across the image. This can be cast as a problem of template matching with every reference symbol as a template. However, the primary challenge associated with such an approach is that the symbols can occur at several scales and might have various degrees of rotation. 
\section{Conclusion}
\label{five}
In this work, a novel framework was proposed to detect and classify various symbols appearing in an engineering drawing. Multiple deep neural networks were employed to detect symbols and classify them. First, two different methods were proposed to detect the table of legends from an engineering drawing. Next, the reference symbols occurring in the table of legends were extracted using simple image processing techniques. A symbol detector was then trained to detect various symbols occurring across the drawing. Lastly, two new methods were proposed to classify the extracted symbols to one of the detected template symbols.





\ifCLASSOPTIONcaptionsoff
  \newpage
\fi

\bibliographystyle{IEEEtran}
\bibliography{main}

\begin{thebibliography}{10}
\providecommand{\url}[1]{#1}
\csname url@samestyle\endcsname
\providecommand{\newblock}{\relax}
\providecommand{\bibinfo}[2]{#2}
\providecommand{\BIBentrySTDinterwordspacing}{\spaceskip=0pt\relax}
\providecommand{\BIBentryALTinterwordstretchfactor}{4}
\providecommand{\BIBentryALTinterwordspacing}{\spaceskip=\fontdimen2\font plus
\BIBentryALTinterwordstretchfactor\fontdimen3\font minus
  \fontdimen4\font\relax}
\providecommand{\BIBforeignlanguage}[2]{{%
\expandafter\ifx\csname l@#1\endcsname\relax
\typeout{** WARNING: IEEEtran.bst: No hyphenation pattern has been}%
\typeout{** loaded for the language `#1'. Using the pattern for}%
\typeout{** the default language instead.}%
\else
\language=\csname l@#1\endcsname
\fi
#2}}
\providecommand{\BIBdecl}{\relax}
\BIBdecl

\bibitem{one}
C.~Moreno-Garc{\'\i}a, E.~Elyan, and C.~Jayne, ``New trends on digitisation of
  complex engineering drawings.'' \emph{Neural Comput. Applic.}, vol.~31,
  no.~6, pp. 1695--1712, 2019.

\bibitem{two}
J.~Nurminen, K.~Rainio, J.~Numminen, T.~Syrj{\"a}nen, N.~Paganus, and
  K.~Honkoila, ``Object detection in design diagrams with machine learning,''
  in \emph{ICCRS}, 2019, pp. 27--36.

\bibitem{three}
E.~Elyan, L.~Jamieson, and A.~Ali-Gombe, ``Deep learning for symbols detection
  and classification in engineering drawings,'' \emph{Neural networks}, vol.
  129, pp. 91--102, 2020.

\bibitem{four}
S.~Mani, M.~Haddad, D.~Constantini, W.~Douhard, Q.~Li, and L.~Poirier,
  ``Automatic digitization of engineering diagrams using deep learning and
  graph search,'' in \emph{CVPRW}, 2020, pp. 176--177.

\bibitem{five}
S.~Barrat, S.~Tabbone, and P.~Nourrissier, ``A {B}ayesian classifier for symbol
  recognition,'' in \emph{GREC}, 2007.

\bibitem{six}
M.~Luqman, T.~Brouard, and J.~Ramel, ``Graphic symbol recognition using graph
  based signature and {B}ayesian network classifier,'' in \emph{ICDAR}, 2009,
  pp. 1325--1329.

\bibitem{seven}
A.~Rezvanifar, M.~Cote, and A.~Albu, ``Symbol spotting for architectural
  drawings: {S}tate-of-the-art and new industry-driven developments,''
  \emph{Trans. Comput. Vis. Appl.}, vol.~11, no.~1, p.~2, 2019.

\bibitem{eight}
A.~Rezvanifar\vspace{0mm}, M.~Cote, and A.~Albu, ``Symbol spotting on digital
  architectural floor plans using a deep learning--based framework,'' in
  \emph{CVPRW}, 2020, pp. 568--569.

\bibitem{nine}
J.~Redmon, S.~Divvala, R.~Girshick, and A.~Farhadi, ``You only look once:
  {U}nified, real-time object detection,'' in \emph{CVPR}, 2016, pp. 779--788.

\bibitem{ten}
S.~Ghosh, P.~Shaw, N.~Das, and K.~Santosh, ``{GSD}--{N}et: {C}ompact network
  for pixel-level graphical symbol detection,'' in \emph{ICDARW}, vol.~1, 2019,
  pp. 68--73.

\bibitem{eleven}
M.~Gori, G.~Monfardini, and F.~Scarselli, ``A new model for learning in graph
  domains,'' in \emph{IJCNN}, vol.~2, 2005, pp. 729--734.

\bibitem{twelve}
G.~Renton, P.~H{\'e}roux, B.~Ga{\"u}z{\`e}re, and S.~Adam, ``Graph neural
  network for symbol detection on document images,'' in \emph{ICDARW}, vol.~1,
  2019, pp. 62--67.

\bibitem{thirteen}
J.~Gilmer, S.~Schoenholz, P.~Riley, O.~Vinyals, and G.~Dahl, ``Neural message
  passing for quantum chemistry,'' in \emph{ICML}, 2017, pp. 1263--1272.

\bibitem{fourteen}
P.~Riba, A.~Dutta, J.~Llad{\'o}s, and A.~Forn{\'e}s, ``Graph--based deep
  learning for graphics classification,'' in \emph{ICDAR}, vol.~2, 2017, pp.
  29--30.

\bibitem{fifteen}
S.~Ren, K.~He, R.~Girshick, and J.~Sun, ``Faster {R--CNN}: {T}owards real-time
  object detection with region proposal networks,'' \emph{NIPS}, vol.~28, pp.
  91--99, 2015.

\bibitem{sixteen}
T.~Lin, M.~Maire, S.~Belongie, J.~Hays, P.~Perona, D.~Ramanan, P.~Doll{\'a}r,
  and C.~Zitnick, ``Microsoft {COCO}: {C}ommon objects in context,'' in
  \emph{ECCV}, 2014, pp. 740--755.

\bibitem{seventeen}
D.~Lowe, ``Distinctive image features from scale--invariant keypoints,''
  \emph{IJCV}, vol.~60, no.~2, pp. 91--110, 2004.

\bibitem{nineteen}
G.~Koch, R.~Zemel, and R.~Salakhutdinov, ``Siamese neural networks for
  one--shot image recognition,'' in \emph{ICMLW}, vol.~2, 2015.

\bibitem{twenty}
F.~Schroff, D.~Kalenichenko, and J.~Philbin, ``Face{N}et: {A} unified embedding
  for face recognition and clustering,'' in \emph{CVPR}, 2015, pp. 815--823.

\bibitem{twentyone}
V.~Nair and G.~Hinton, ``Rectified linear units improve restricted {B}oltzmann
  machines,'' in \emph{ICML}, 2010.

\end{thebibliography}

\end{document}